\documentclass{article} 
\usepackage{fullpage}

\usepackage{wrapfig,lipsum,booktabs}
\usepackage[utf8]{inputenc} 
\usepackage[T1]{fontenc}    
\usepackage{url}            
\usepackage{booktabs}       
\usepackage{amsfonts}       
\usepackage{nicefrac}       
\usepackage{microtype}      
\usepackage{soul}
\usepackage{graphicx}
\usepackage{amsmath}
\usepackage{outlines}
\usepackage{xurl}

\usepackage[colorlinks=true,
citecolor=blue,
filecolor=black,
linkcolor=blue,
urlcolor=blue]{hyperref}

\usepackage{amsmath,amsfonts,bm}

\def\eqref#1{equation~\ref{#1}}

\def\1{\bm{1}}

\DeclareMathAlphabet{\mathsfit}{\encodingdefault}{\sfdefault}{m}{sl}
\SetMathAlphabet{\mathsfit}{bold}{\encodingdefault}{\sfdefault}{bx}{n}

\usepackage{multirow}
\usepackage{paralist}

\usepackage{multicol}
\usepackage{diagbox}

\usepackage[ruled,noend]{algorithm2e}

\SetCommentSty{mycommfont}

\usepackage{here}

\usepackage{amsmath,amssymb,amsfonts,amsbsy,amsfonts,latexsym}
\usepackage{makecell}
\usepackage{xcolor}
\usepackage{colortbl}

\usepackage{tabularx,colortbl,xcolor}
\usepackage[normalem]{ulem}
\useunder{\uline}{\ul}{}

\usepackage{enumitem}

\usepackage{xparse}

\SetKwInput{KwInput}{Input}
\SetKwInput{KwRequire}{Require}

\usepackage{longtable}

\NewDocumentCommand{\var}{O{s} m O{}}{%
  \ensuremath{#1_{#2}^{#3}}
}
\usepackage{siunitx}

\newcommand{\commentout}[1]{}

\definecolor{light-gray}{gray}{0.80}

\newcommand\fref{Figure~\ref}
\newcommand\tref{Table~\ref}
\newcommand\sref{Section~\ref}

\newcommand{\red}[1]{{\color{red} #1}}

\usepackage{amsthm}

\usepackage{amsthm}

\newtheorem{mytheorem}{Theorem}
\newtheorem{assumption}[mytheorem]{Assumption}
\newtheorem{lemma}[mytheorem]{Lemma}
\newtheorem{remk}[mytheorem]{Remark}

\usepackage{subfig}
\newcommand{\llms}{LLMs\xspace}
\newcommand{\lvlms}{LVLMs\xspace}
\newcommand{\qwen}{QWen-VL\xspace}
\newcommand{\sparkles}{SparklesChat\xspace}
\newcommand{\llava}{LLaVA\xspace}
\newcommand{\otter}{Otter\xspace}

\newcommand{\ca}{CA\xspace}
\newcommand{\cra}{CrA\xspace}
\newcommand{\mmca}{MMCA\xspace}
\newcommand{\llama}{LLaMa-2\xspace}
\newcommand{\ourssingle}{DeepSpeed-VisualChat-Single\xspace}
\newcommand{\ours}{DeepSpeed-VisualChat\xspace}
\usepackage{pifont}

\usepackage{arydshln}
\usepackage{xcolor}
\usepackage{listings}
\definecolor{myblue}{rgb}{0,0.2,0.8}
\hypersetup{ %
pdftitle={},
pdfauthor={},
pdfsubject={},
pdfkeywords={},
pdfborder=0 0 0,
pdfpagemode=UseNone,
colorlinks=true,
linkcolor=myblue,
citecolor=myblue,
filecolor=myblue,
urlcolor=myblue,
pdfview=FitH}
\makeatletter
\def\adl@drawiv#1#2#3{%
        \hskip.5\tabcolsep
        \xleaders#3{#2.5\@tempdimb #1{1}#2.5\@tempdimb}%
                #2\z@ plus1fil minus1fil\relax
        \hskip.5\tabcolsep}
\newcommand{\cdashlinelr}[1]{%
  \noalign{\vskip\aboverulesep
           \global\let\@dashdrawstore\adl@draw
           \global\let\adl@draw\adl@drawiv}
  \cdashline{#1}
  \noalign{\global\let\adl@draw\@dashdrawstore
           \vskip\belowrulesep}}
\makeatother

\definecolor{upforestgreen}{rgb}{0.6, 0.8, 0.2}

\usepackage{chngcntr}
\usepackage{adjustbox}
\usepackage{wrapfig}
\usepackage{arydshln}

\usepackage{tcolorbox}

\begin{document}

\title{\ours: Multi Round Multi Images Interleave Chat via Multi-Modal Casual Attention}

\author{
 Zhewei Yao, Xiaoxia Wu, Conglong Li, Minjia Zhang, Heyang Qin\\
 Olatunji Ruwase, Ammar Ahmad Awan, Samyam Rajbhandari, Yuxiong He
 \\
\\  DeepSpeed of Microsoft 
}

\date{}
\maketitle

\begin{abstract}


Most of the existing multi-modal models, hindered by their incapacity to adeptly manage interleaved image-and-text inputs in multi-image, multi-round dialogues, face substantial constraints in resource allocation for training and data accessibility, impacting their adaptability and scalability across varied interaction realms. 
To address this, we present the \ours framework, designed to optimize Large Language Models (LLMs) by incorporating multi-modal capabilities, with a focus on enhancing the proficiency of Large Vision and Language Models in handling interleaved inputs. 
Our framework is notable for (1) its open-source support for multi-round and multi-image dialogues, (2) introducing an innovative multi-modal casual attention mechanism, and (3) utilizing data blending techniques on existing datasets to assure seamless interactions in multi-round, multi-image conversations. 
Compared to existing frameworks, \ours shows superior scalability up to 70B parameter language model size, representing a significant advancement in multi-modal language models and setting a solid foundation for future explorations.\footnote{Code will be released soon as a part of \url{https://github.com/microsoft/DeepSpeedExample}}

\end{abstract}

\section{Introduction}
\label{sec:introduction}
State-of-the-art large language models (\llms) like GPT~\cite{brown2020language,OpenAI2023GPT4TR} have showcased exceptional prowess in myriad text generation and comprehension tasks, especially when subjected to zero-/few-shot learning. Once these models undergo supervised fine-tuning or reinforcement learning combined with human feedback, their proficiency in versatile interactive challenges—ranging from coding tasks~\cite{copilot} to quantitative reasoning \cite{lewkowycz2022solving}, mathematical proofs~\cite{imani-etal-2023-mathprompter,welleck2022naturalprover}, and AI chatbot interactions~\cite{chatgpt,chatllama,ouyang2022training,yao2023dschat}—becomes comparable to human experts.


Seeking to transcend the bounds of text-only processing inherent to \llms, numerous researchers have made strides in endowing these models with multi-modal capabilities. These advances span across various modalities such as images, audios, and videos, often achieved via feature alignment and model alterations~\cite{girdhar2023imagebind, gong2023multimodal,zhu2023minigpt,liu2023visual,li2023otter, bai2023qwen,huang2023sparkles}. Notably, among these multi-modal endeavors, large vision and language models (\lvlms) have garnered significant interest~\cite{zhu2023minigpt, liu2023visual}, mainly owing to their potential in facilitating comprehensive visual-textual understanding.



Current frameworks and studies largely focus on either (1) tasks related to individual images, like visual question answering and captioning~\cite{liu2023visual}, or (2) handling multiple images but requiring concurrent input~\cite{li2023otter}. Neither approach adeptly manages interleaved image-and-text inputs. The \qwen framework~\cite{bai2023qwen}, an extension of the \llava architecture~\cite{liu2023visual}, makes progress in this direction. However, its training costs prove prohibitive for many research labs, and it withholds its training data.  In parallel, the \sparkles model~\cite{huang2023sparkles}, annotated by GPT4, relies on continuous training with MiniGPT4~\cite{zhu2023minigpt} due to its limited dataset. Both \qwen and \sparkles adopt prevailing \lvlms designs without innovative architectural exploration. Additionally, in multi-image contexts, their performance is found lacking, even with significant training investments~\footnote{We count the pretraining cost of MiniGPT4 as a part of~\sparkles}, as shown in our comparisons~\fref{tab:compare1}.


While larger language models typically demonstrate superior generation abilities compared to their smaller counterparts, prevailing frameworks~\cite{li2023otter,liu2023visual,zhu2023minigpt,bai2023qwen} predominantly concentrate their efforts on \llms with 7 or 13 billion parameters (e.g., \llama-7B or \llama-13B~\cite{touvron2023llama}). This focus restricts the exploration of the extensive capabilities inherent to larger \llms.


\begin{figure}
    \centering
    \includegraphics[width=0.83\textwidth]{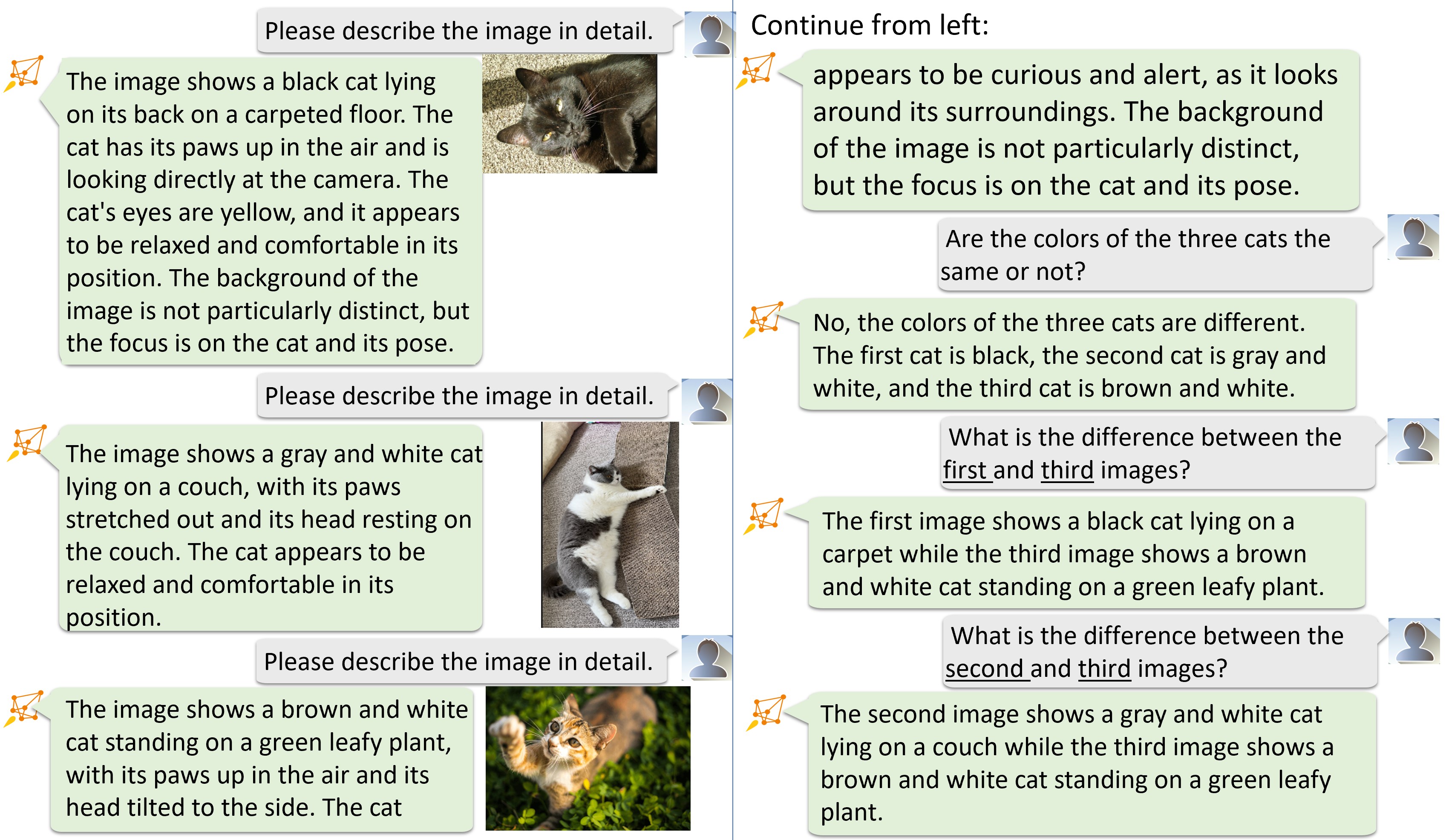}
\caption{An example of \ours.
}
\label{fig:banner}
\end{figure}

To address the aforementioned challenges, we introduce the DeepSpeed Multi Round and Multi Images Chat framework (\ours), offering several key contributions:
\begin{itemize}
    \item \textbf{Fully Open-Sourced Multi-round Multi-image Framework:} \ours, one of the pioneering fully open-sourced frameworks, enables multi-round and multi-image dialogues, accommodating interleaved text-and-image inputs, as visualized in~\fref{fig:banner}. 
    
    \item \textbf{Multi-Modal Casual Attention (MMCA):} We devise a novel \mmca for multi-modal models that independently computes attention weights across various modalities. \mmca attains objectives analogous to conventional cross-attention mechanisms~\cite{li2023otter}, yet offers enhanced casual attention interpretations for generative tasks, eliminating the need for additional modules or parameters, and it presents superior training data efficiency compared to standard casual attention~\cite{zhu2023minigpt,liu2023visual}.
    \item \textbf{Data Blending for Interleaved Inputs: }  To facilitate conversations with interleaved modalities, \ours employs assorted data blending techniques on existing datasets, overcoming the shortage of interleaved text-and-image inputs in most available open-sourced datasets.
    
    \item \textbf{Unprecedented Scalability:} We leverage the DeepSpeed framework~\cite{rasley2020deepspeed} to amplify our training with a 2B visual encoder from~\cite{ilharco_gabriel_2021_5143773} and a 70B language decoder from \llama~\cite{touvron2023llama}, illustrating the remarkable scalability of our framework.
    
\end{itemize}
These innovations demonstrate our commitment to progressing multi-modal conversational AI models, ensuring enhanced interoperability, attention mechanism sophistication, scalable solutions, and comprehensive dataset utilization.
\section{Related Work}
\label{sec:related_work}

Multi-modal models, especially those focusing on vision-language integrations, typically fall into two distinct categories: dual-encoder-based models~\cite{lxmert,vilbert,vl-bert,vinvl,clip,oscar,vilt,albef,simvlm,vlmo,ofa,coca,vlbeit,blip,pali,clip,align,vilclip,klite,florence,flava,liu2023learning,shen2023scaling}, and models comprising visual encoders and textual decoders~\cite{flamingo,li2023blip2,liu2023visual,li2023otter,zhu2023minigpt,gong2023multimodal,su2023pandagpt,bai2023qwen,awadalla2023openflamingo,huang2023sparkles}. Our work is associated with the latter, often referred to as Large Visual Language Models (\lvlms), thus our discussion predominantly revolves around \lvlms.


Most implementations of \lvlms deploy one of two architecture styles: (1) The Flamingo design~\cite{flamingo,li2023otter,awadalla2023openflamingo} incorporates cross-attention, introducing new parameters to \llms to interlink visual and textual elements. (2) The Flamingo design~\cite{flamingo,li2023otter,awadalla2023openflamingo} incorporates cross-attention, introducing new parameters to \llms to interlink visual and textual elements. Although both designs effectively assimilate visual information and generate textual content, their advantages and drawbacks are manifold. The Flamingo design necessitates extensive training/inference memory and fewer data due to the introduction of numerous new parameters. Conversely, the MiniGPT4 design, while less memory-intensive, is more data-dependent to effectively align visual and textual features. Consequently, an emerging query is whether a novel architecture can harmonize the introduction of fewer new parameters with data efficiency.



Despite the substantial advancements achieved by existing \lvlms, certain aspects, particularly multi-round multi-image conversation involving interleaved image and text input, remain unaddressed. Works like~\cite{liu2023visual} predominantly concentrate on single image conversation, and~\cite{li2023otter} necessitate simultaneous input of all images, limiting their applicability in conventional conversational applications.

The paucity of data pertinent to these scenarios has led researchers to explore available data to facilitate new applications. A contemporary work, \sparkles~\cite{huang2023sparkles}, exploits the GPT-4~\cite{gpt4} API to synthesize several thousands of multi-round multi-image data in a unique format. However, \sparkles does not innovate any new architectural modifications and relies on the pre-trained model, MiniGPT4~\cite{zhu2023minigpt}, thereby incurring additional training costs and complexity. Moreover, \sparkles does not exhibit any superior performance in unseen multi-image tasks compared to \ours, even without utilizing multi-image interleaved chat data. Refer to~\fref{tab:compare1} and~\ref{tab:compare3} for detailed examples.

\begin{wrapfigure}{r}{0.45\textwidth}
    \centering
    \vspace{-1.6cm}
    \includegraphics[width=0.45\textwidth]{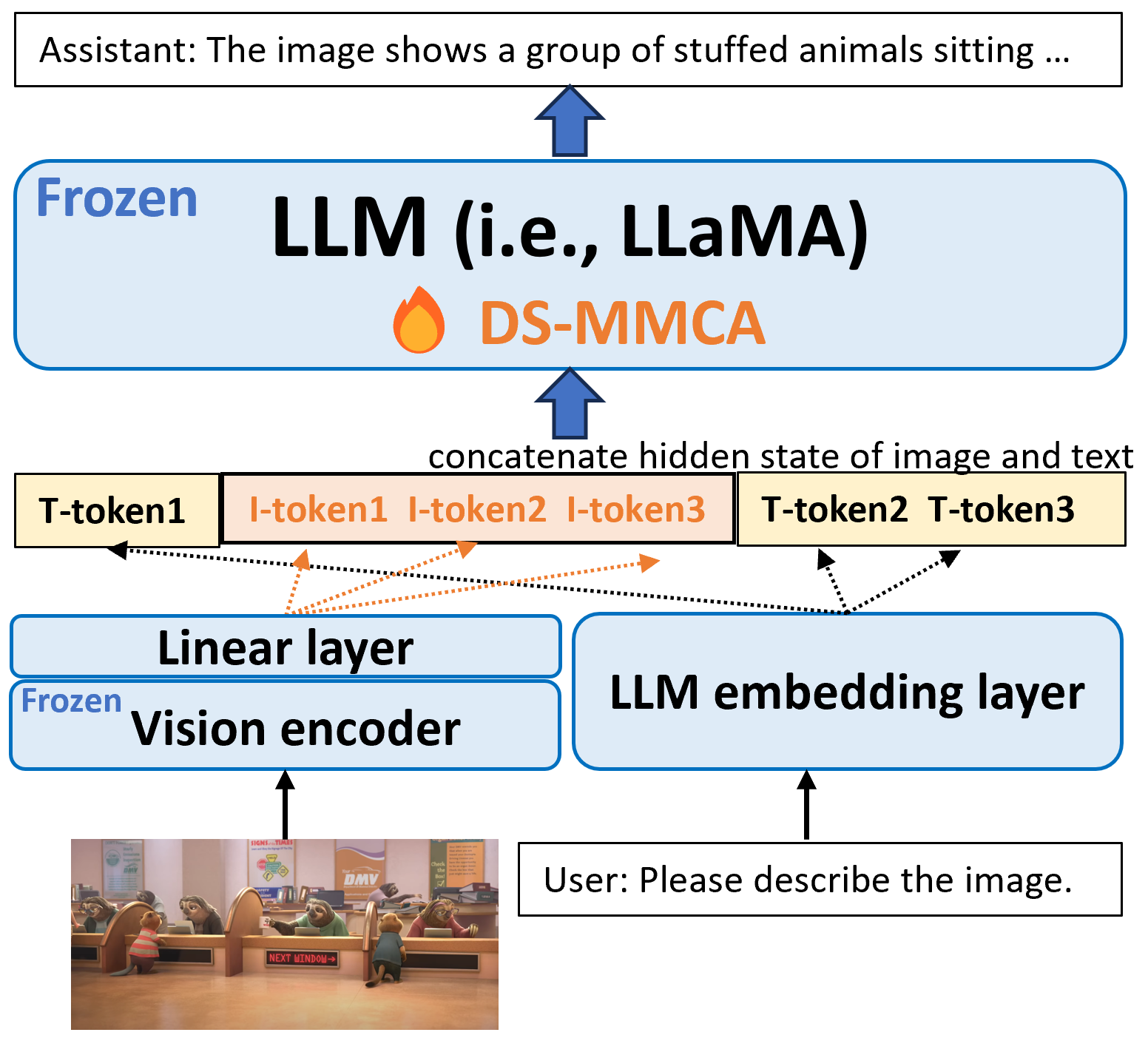}
\caption{Model Structure. A pre-trained vision encoder encodes an image which is then projected through a linear layer to align with the hidden dimension of the text embedding layer's output. These different inputs are subsequently merged and forwarded to language models like \llama powered by our new Multi-Modal Casual Attention (MMCA) mechanism. Here, both the vision encoder and the language model are frozen.
}
\label{fig:model}
 \vspace{-1.3cm}
\end{wrapfigure}

\section{Method}
Our model architecture is built on the structure of MiniGPT4~\cite{zhu2023minigpt,liu2023visual}, as depicted in~\fref{fig:model}. Specifically, we maintain the entirety of the visual encoder and the whole language model, with the exception of the embedding layer, in a frozen state.
Thus, the only trainable parameters within our model are the visual feature projection layer (a linear layer) and the language model's embedding.
In total, our set of trainable parameters ranges around $O(10M)$ to $O(100M)$, primarily contingent on the size of the embedding layer.

Diverging from the previous MiniGPT4 architecture, we substitute the conventional casual attention mechanism with our proposed multi-modal casual attention mechanism (refer to~\sref{sec:mmca}). This modification solely alters the computation of casual attention and does not incorporate any new parameters.
 

We adopt a unified instruction tuning format for all experiments, and the template is shown in~\fref{fig:ift_template}.
It is crucial to note that we do not employ any special tokens as our prefix; for example, “\#\#\# Image i” is not a special token, and the tokenizer interprets it as a regular string.

\begin{figure}
\center{ 
\begin{lstlisting}
<System Insturction>       % You are a powerful vision-language assistant.

### Image 1: <image>       % some image, e.g., cat-1.png
### Question: <question>   % please describe the image.
### Answer: <answer>       % It's a cute black cat.

### Image 2: <image>       % some image, e.g., cat-2.png
### Image 3: <image>       % some image, e.g., cat-3.png
### Question: <question>   % What's difference between three cats?
### Answer: <answer>       % The color of three cats are different. 

...
\end{lstlisting}
}
\caption{Here <System Instruction>, <question>, <answer> can be simply replaced by text, and <image> can be replaced by real image tokens. 
The content after ``\%'' is an example.}
\label{fig:ift_template}
\end{figure}

In alignment with the recent trend of instruction fine-tuning, the final loss of our model is calculated solely on “<answer>”, as illustrated in~\fref{fig:ift_template}. If multiple conversations are present, we compute the loss for all corresponding “<answer>” instances.

Throughout the paper, unless specifically mentioned, we employ the \llama family as our language and utilize the extracted (and frozen) visual encoder from \qwen~\cite{bai2023qwen} as our visual encoder, which accepts 448x448 images and produces 256 image tokens per image. The rationale for opting for \qwen’s encoder over the typically utilized CLIP~\cite{clip} is elaborated in~\sref{sec:si_learning}. The sequence length for training \llama is capped at 4096. When referring to our model as \ours-xB (e.g., \ours-13B), the size is exclusively contingent on the language model components (\llama-13B).




\section{Multi-Round Single-Image Exploration}
\label{sec:multi-round-single-image}

\subsection{Multi-Modal Casual Attention}
\label{sec:mmca}

There are two common attention mechanisms used to connect the visual and textual components in a multi-modal model: causal attention, as used in~\cite{zhu2023minigpt, bai2023qwen}, and cross attention, as used in~\cite{li2023otter, flamingo}.

\begin{figure}
    \centering
    \includegraphics[width=1.03\textwidth]{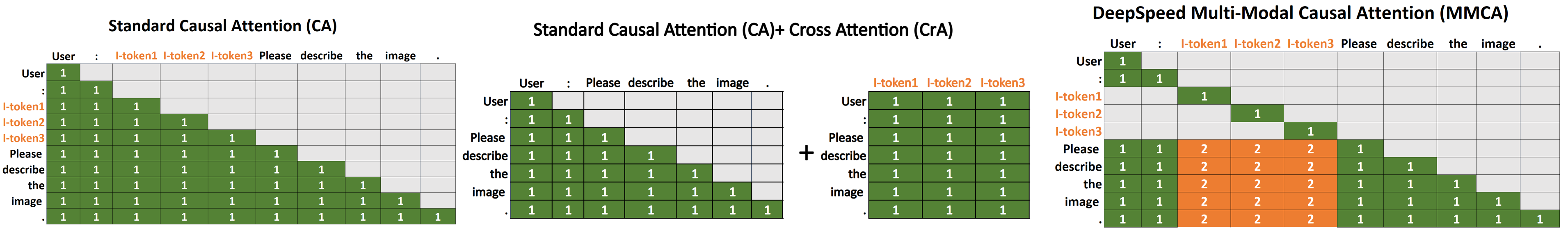}
\caption{Different Attention Mechanisms: Examine the differing attention mechanisms using an input sentence "User: Please describe the image." coupled with three \textbf{Image} tokens (I-token1, I-token2, I-token3). On the left, we demonstrate standard causal attention~\cite{zhu2023minigpt, bai2023qwen}, treating image tokens as text. In the middle, we present cross attention applied to images, while maintaining standard causal attention for text tokens. On the right, we illustrate our innovative multi-modal attention proposal where image tokens only perform self-attention, and text tokens attend to text/image tokens independently, highlighted with an orange mask. This mechanism is defined by: $\text{softmax}(QK^T \odot M_1)+\text{softmax}(QK^T\odot M_2)$ with $Q$ and $K$ as query and key, $M_1=[M==1]$, and $M_2=[M==2]$, with $M\in\mathbf{R}^{10\times10}$ in this case.
}
\label{fig:attention}
\end{figure}


\textbf{Causal Attention (\ca)}: The \ca-based method simply projects visual features (i.e., the features from the output of the final visual encoder layer) into textual features and combines them with the normal textual features after the textual embedding layer to feed into \llms. The benefit of \ca is that it's a natural extension of the original attention mechanism in \llms, and as such, it doesn't introduce any extra modules or parameters. However, this approach raises some intuitive problems:

(1) For a visual token, it attends to previous visual and textual tokens, even though visual tokens are already fully encoded in a bidirectional manner and don't need further attention from other visual tokens or the beginning of textual tokens.

(2) For a textual token, it needs to learn how to distribute its attention weights between its previous textual and image tokens. Due to these issues, we found that the data efficiency of \ca in \lvlms is often problematic. To address this, \llava and \qwen require visual-language pretraining to fully align visual features with textual features. We also test and compare it with our proposed \mmca in~\sref{sec:result_of_mrsi}.

\textbf{Cross Attention (\cra)}: The alternative, cross attention (\cra), along with \ca, exhibits better data efficiency but also comes with a few drawbacks:

(1) It introduces new parameters to the model. For example, \otter has more than 1.5 billion trained parameters compared to the millions of trained parameters in \llava. This significantly increases the training cost and memory requirements.

(2) It requires careful design if an image is introduced in the middle of a conversation during training, as previous text tokens should not be able to attend to the image.




\textbf{Multi-Modal Causal Attention Mechanism (\mmca)}: To overcome these issues, we propose a new multi-modal causal attention mechanism (\mmca). The overall idea is as follows:

(1) For visual tokens, they only attend to themselves, as visual tokens are  encoded by the visual encoder.

(2) For textual tokens, they attend to all their previous tokens. However, they have two separate attention weight matrices for their previous textual tokens and image tokens.

The intuition behind the second point of \mmca is that the attention weight for one modality may affect the other modality. For instance, a textual token may pay more attention to textual information than visual information. Therefore, if the attention weight matrix is normalized across both modalities, the attention score for visual tokens might be very small. Refer to \fref{fig:attention} for a visualization of the three attention mechanisms.

\subsection{Result}
\label{sec:result_of_mrsi}
\subsubsection{Comparison between Different Attentions}
\label{sec:comparison_between_attn}

\paragraph{Experimental Setting}
We employ the \llama-7B language model in conjunction with the \qwen-visual-encoder as our visual encoder. These two models are connected via a straightforward linear projection layer. Our model underwent training on two LLaVa datasets, as outlined in the initial two rows of~\tref{tab:data-summary}.

During training, all models were run for 5 epochs with a training batch size of 128. Our primary evaluation focused on single-image captioning and single-image Visual Question Answering (VQA). The peak learning rate was set to 1e-3 for both the projection linear layer and the embedding layer, and we employed the AdamW optimizer~\cite{loshchilov2017decoupled} with first- and second-order coefficients set to (0.0, 0.95).

For dataset splitting, we divided the training and validation datasets in a 90/10 ratio across the entire dataset. Additionally, we incorporated 10\% of the total training iterations as warm-up steps. Our training framework of choice was DeepSpeed~\cite{rasley2020deepspeed}, and we utilized FP16 training to expedite the training process.
\begin{table}[t]
\centering
\footnotesize
\caption{Training datasets summary. Due to context length limitation, for otter\_mimicit\_sn, otter\_mimicit\_tvc, and otter\_mimicit\_vst datasets we only used the samples with $\leq$ 8 images.}\label{tab:data-summary}
\begin{tabular}{@{}lrl@{}}
\toprule
Name & Num. samples & Description \\
\midrule
(1) llava & 49924 & The detail description and complex reasoning data used by the \llava model~\cite{liu2023visual}. \\
 &  & Randomly concatenate 1 to 3 samples into one sample. Details in~\sref{sec:data-blending}. \\
(2) llava\_dial & 37818 & The conversation data used by the \llava model~\cite{liu2023visual}. \\
 &  & Randomly concatenate 1 to 2 samples into one sample. Details in~\sref{sec:data-blending}. \\
(3) otter\_mimicit\_cgd & 70940 & The COCO (General) data used by the \otter model~\cite{li2023otter}. \\
(4) otter\_mimicit\_sd & 8006 & The SD (Surveillance) data used by the \otter model~\cite{li2023otter}. \\
(5) otter\_mimicit\_sn & 487 & The SN (Indoor Ego.) data used by the \otter model~\cite{li2023otter}. \\
(6) otter\_mimicit\_tvc & 2 & The TVC (TV) data used by the \otter model~\cite{li2023otter}. \\
(7) otter\_mimicit\_vst & 115 & The VIST (Story) data used by the \otter model~\cite{li2023otter}. \\
(8) llava\_otter\_blend & 48869 & Data blended from llava, llava\_dial, otter\_mimicit\_cgd. Details in~\sref{sec:data-blending}. \\
(9) sparkles\_dialogue & 6520 & The SparklesDialogue data used by the \sparkles model~\cite{huang2023sparkles}. \\
\midrule
Total & 222681 & \\
\bottomrule
\end{tabular}
\vspace{-0.2cm}
\end{table}

Throughout this work, we mainly compare the generation capability of different models on certain examples without comprehensively testing models on existing benchmark. 
Please see more details in~\sref{sec:conclusions} for limitations of our work.

\begin{figure}
    \centering
    \includegraphics[width=0.8\textwidth]{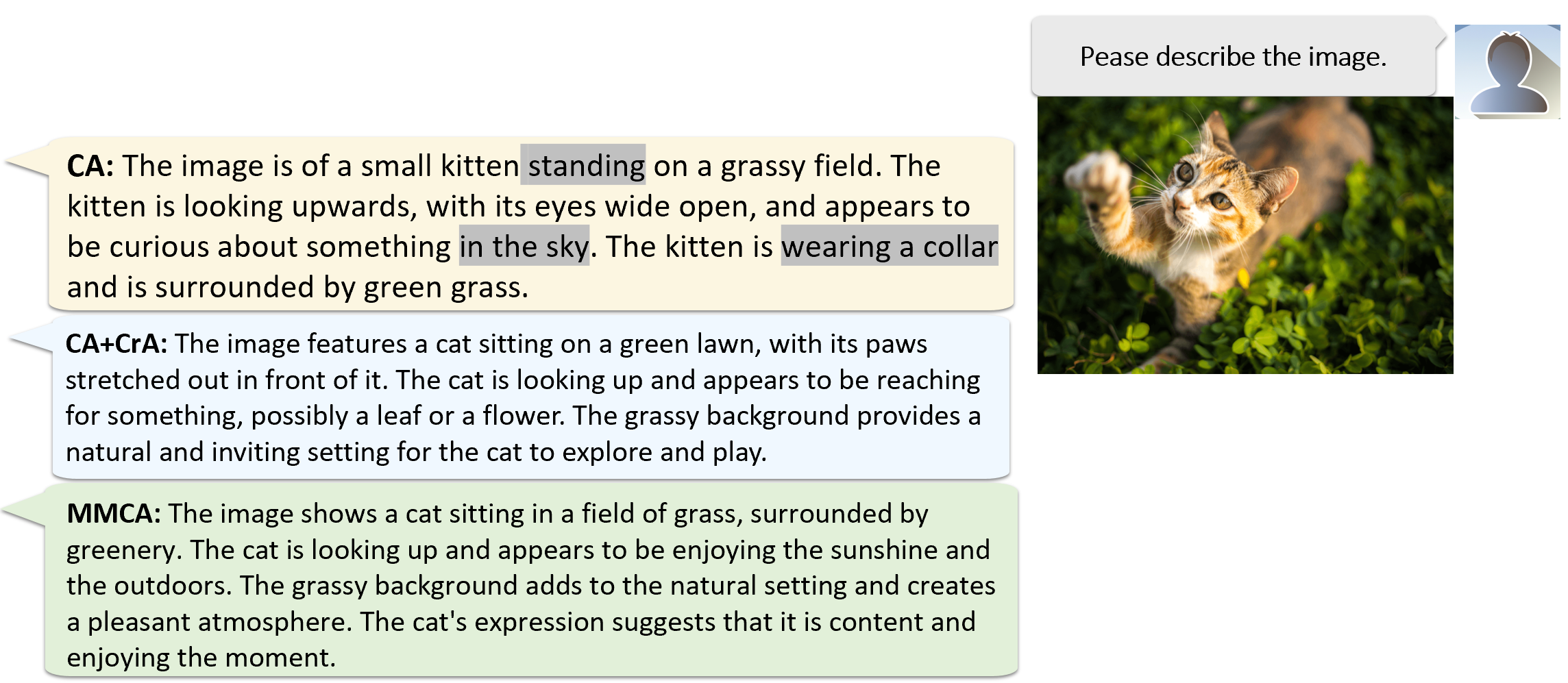}
\caption{
Example visual and language inputs that demonstrate the output comparison between (1) the standard causal attention (\textbf{\ca}) (2)  the standard causal attention combined with cross-attention (\textbf{\ca+ \cra}) and (3) the special multi-modal causal attention (\textbf{\mmca}) in \ourssingle. 
}\label{tab:attention-comparison-cat}
\end{figure}

\begin{figure}
    \centering
    \includegraphics[width=0.8\textwidth]{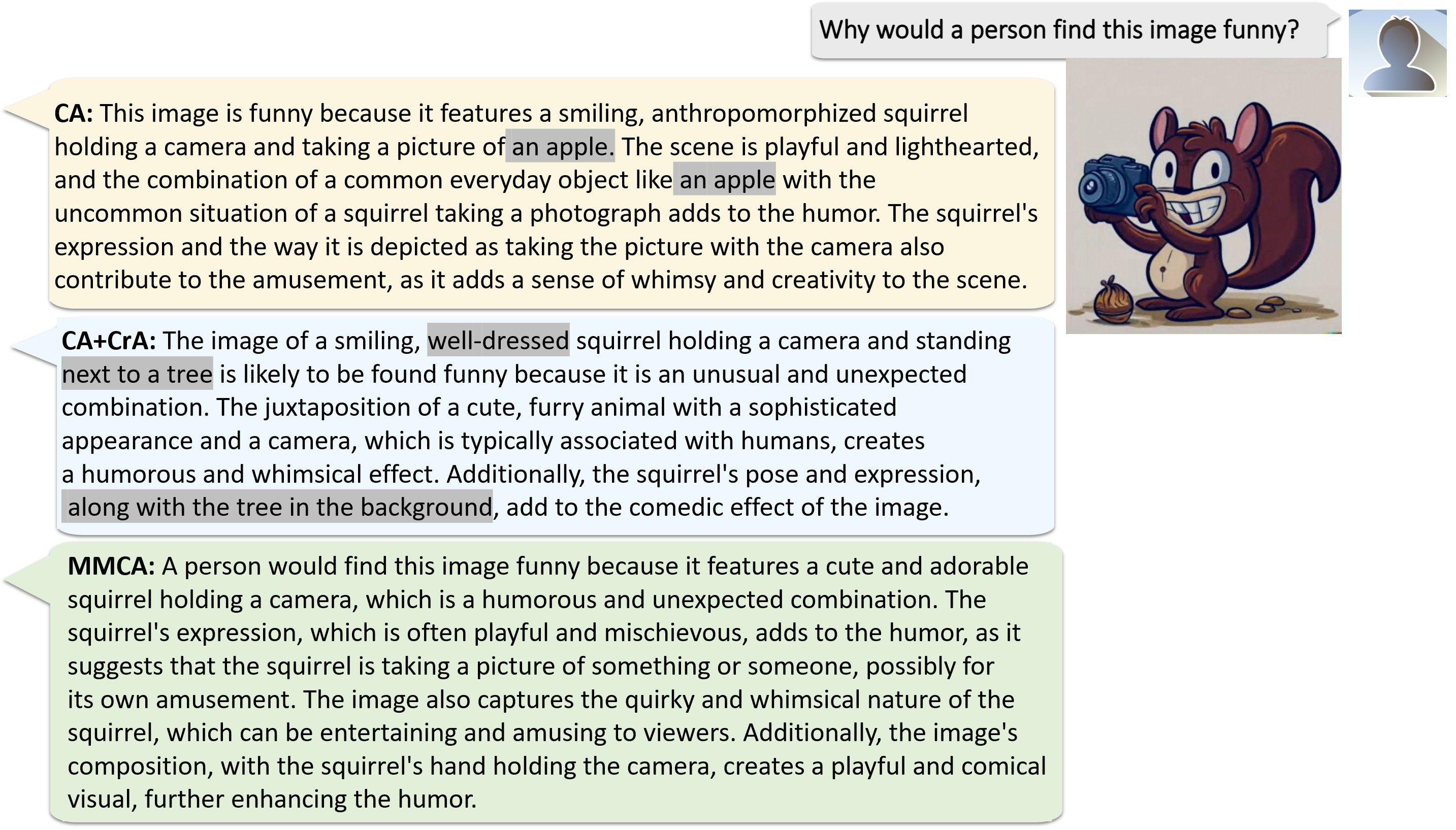}
\caption{
\ourssingle accurately identifies the squirrel and camera in the image, while the baseline model mistakenly includes “standing next to a tree”. 
}
\label{tab:attention-comparison-quirrel}
\end{figure}

\begin{figure}
    \centering
    \includegraphics[width=0.8\textwidth]{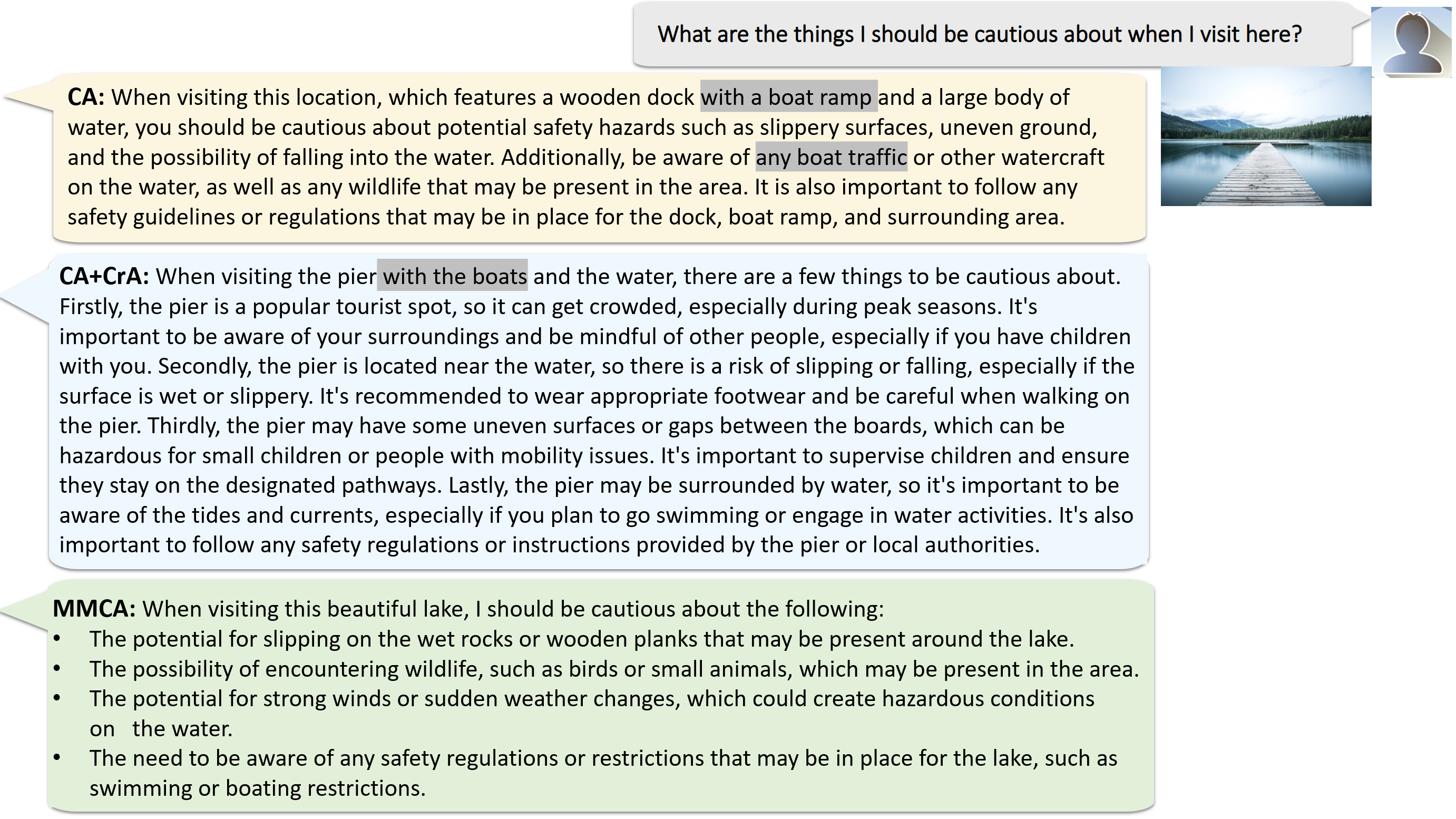}
\caption{
\ourssingle accurately identifies the scene as a beautiful lake and offers a set of plausible suggestions. In contrast, the baseline misinterprets the image as containing “dock with a boat ramp”. } \label{tab:attention-comparison-lake}
\end{figure}

\paragraph{Demo results.} We begin by showcasing various examples that highlight the capabilities of \ours in single-image visual language conversations, employing different attention mechanisms. As demonstrated in~\fref{tab:attention-comparison-cat}, \fref{tab:attention-comparison-quirrel}, and~\fref{tab:attention-comparison-lake}, \ours, when coupled with \mmca, effectively discerns visual details in images and furnishes coherent responses to user queries. 

Furthermore, \ours exhibits a more comprehensive and precise grasp of image details compared to alternative attention mechanisms, such as the use of combined masks from both causal attention and cross attention. It is also evident that, in contrast to the combination of \cra and \ca, as well as \mmca, \ca alone may exhibit slightly more errors (\fref{tab:attention-comparison-cat}) and capture a lower degree of reasoning capability (\fref{tab:attention-comparison-lake}).

\subsubsection{Result of \ourssingle}
\paragraph{Experimental Setting}
All settings remain consistent with those outlined in~\sref{sec:comparison_between_attn}, with the exception of two modifications: an increase in the language model size from \llama-7B to \llama-13B and an extension of the training epoch count from 5 to 10.
\paragraph{Demo results}
Upon elevating the language model size from 7B to 13B, we observe a marked improvement in the model's ability to recognize images and capture the logical connections between questions and images.

\begin{figure}
    \centering
    \includegraphics[width=0.8\textwidth]{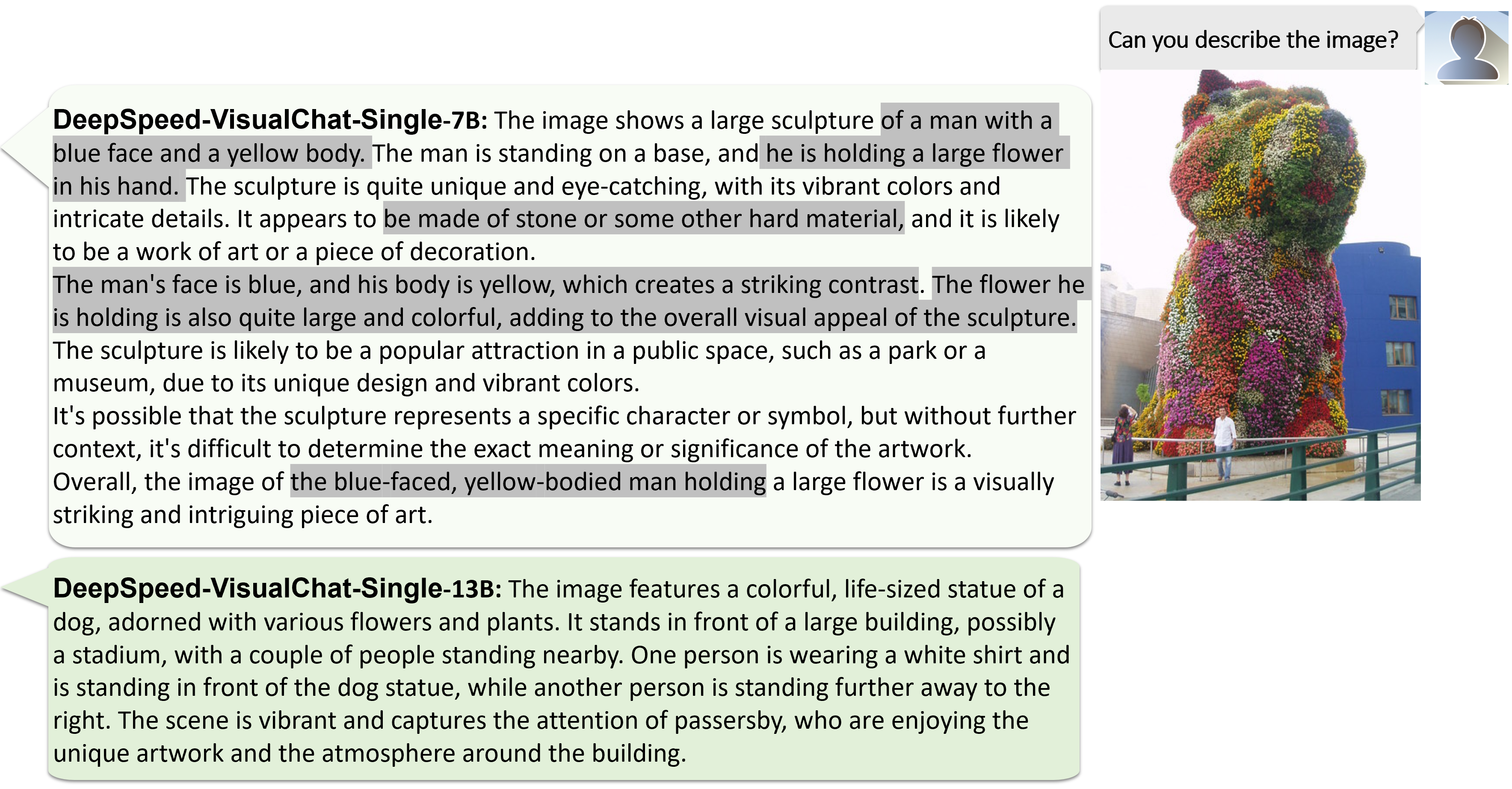}
    \centering
    \includegraphics[width=0.8\textwidth]{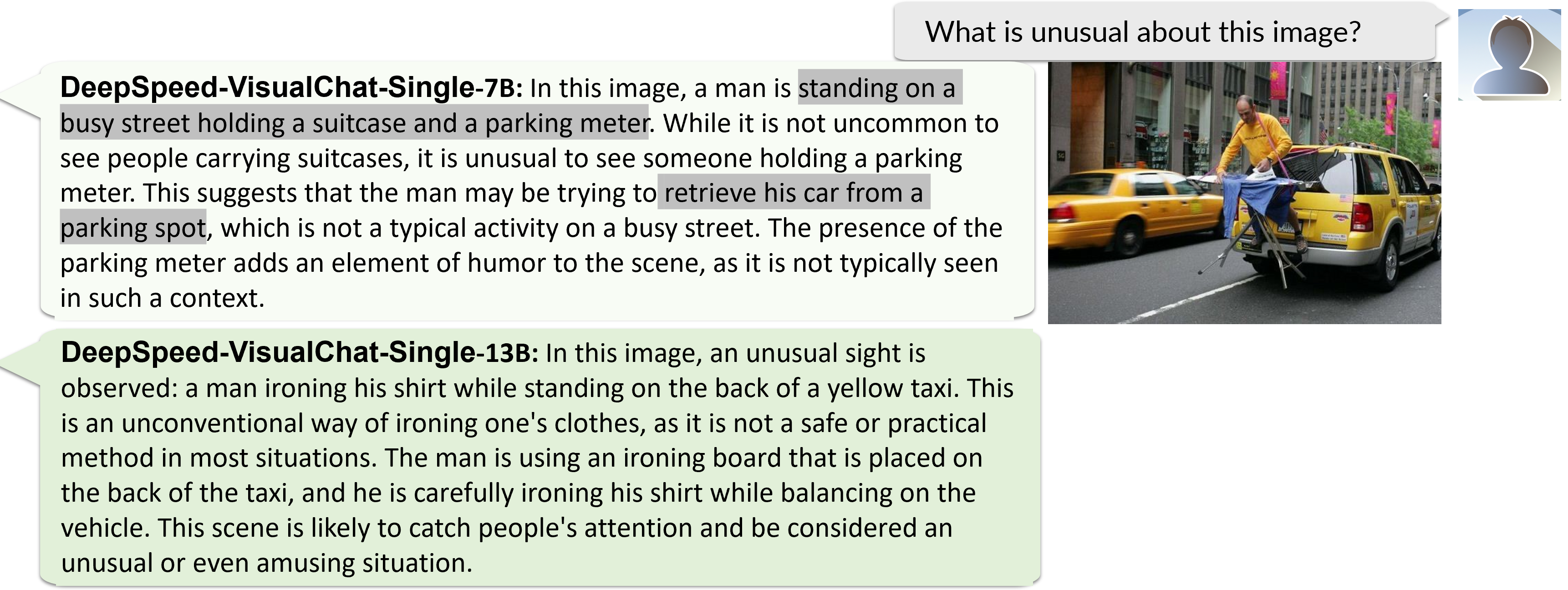}
\caption{
The above two examples illustrate the difference between \ourssingle-13B and \ourssingle-7B.
}
\label{tab:7b-13b-scupture}
\end{figure}

Additionally, referring to the upper example in~\fref{tab:7b-13b-scupture}, aside from the issue of incorrect descriptions, it becomes apparent that \ourssingle-7B occasionally exhibits repetitive context. However, this problem is less frequent with \ourssingle-13B, owing to the enhanced generative capabilities of larger language models. This observation aligns with the superior performance typically associated with larger language models.

\subsection{Other Learning}
\label{sec:si_learning}
Throughout the training process of \ourssingle, we accumulated several additional lessons. It's important to note that most of these observations lack sufficient evidence and require further exploration. We present them here to assist others, but they should not be considered final conclusions.
\begin{itemize}
    \item \textbf{Better Visual Encoder:}  Commonly, the CLIP visual encoder is used in \lvlms. However, the CLIP encoder's resolution is limited to 224x224, which restricts the level of detail in the images. In our testing, we discovered that using the newly released visual encoder from \qwen significantly improves the final model quality due to its higher input resolution (448x448) and larger encoder size (2B parameters).
    \item \textbf{Overfitting or Not:} Typically, we select the best evaluation checkpoint or one close to it for final testing. However, during \ourssingle training, we found that the final checkpoint, even if it appears overfitted, often delivers better testing results compared to middle checkpoints. Does this imply that we should intentionally overfit our model? The answer is no. We experimented with 5, 10, and 20 epochs for \ourssingle-13B and observed that 10-epoch training typically yields superior final model quality.
    
    \item \textbf{Adding LoRA to Visual Encoder or Language Decoder:} We attempted to introduce LoRA-based training to enhance model quality. However, applying LoRA to either module did not yield any significant  benefits.
    
    \item \textbf{Lowering the Learning Rate for Pretrained Components:} We experimented with a smaller learning rate for language embedding since it is already pretrained. However, our results indicated that there is no significant difference when using a separate lower learning rate.
    
    \item \textbf{Using Chat-/Non-Chat-Based Models:} We explored both chat-based and non-chat-based LLama-2 models. Our findings suggest that when using the chat-based model, strict adherence to the chat-based model's instruction tuning format is crucial. Failing to do so resulted in even worse model quality than the non-chat-based model.
    
    \item \textbf{Inserting New Special Tokens or Not:}  As illustrated in~\fref{fig:ift_template}, a few tokens can be replaced by new inserted special tokens, such as encoding "\#\#\#Human: " as a new special token. However, our testing revealed that it is better not to incorporate them as special tokens. Introducing them as special tokens significantly worsened our generation performance compared to the previous approach.

\end{itemize}

\section{Multi-Round Multi-Image Exploration}
\label{sec:multi-round-multi-image}
\subsection{Data Blending}
\label{sec:data-blending}

One critical missing element for enabling multi-round and multi-image conversations is data. The sole source of multi-round multi-image data we located is the SparklesDialogue dataset~\cite{huang2023sparkles}, which contains a mere 6520 samples. To address this limitation, we employed two methods to synthesize multi-round multi-image data from existing single-image or single-round data: simple data concatenation and \llava-\otter data blending.

\subsubsection{Simple data concatenation}
For the "llava" and "llava\_dial" datasets utilized by the \llava model, each sample comprises single/multi-round conversations for a single image. To simulate scenarios where a user sequentially asks questions about multiple images, we conducted straightforward data post-processing for these two datasets. Specifically, we randomly concatenated different numbers of samples into a single sample. In the case of "llava," we concatenated 1 to 3 samples, while for "llava\_dial," we concatenated 1 to 2 samples (refer to \tref{tab:data-summary}).

\subsubsection{\llava-\otter data blending}
We noticed that the llava and llava\_dial datasets used by \llava model and the otter\_mimicit\_cgd dataset used by the \otter model all use the COCO train2017 images. For the llava and llava\_dial datasets, each sample includes a single/multi-round conversations for a single image. For the otter\_mimicit\_cgd dataset, each sample includes a single-round conversation for a pair of images. This enables us to build a synthesized multi-round multi-image data llava\_otter\_blend as a more natural blending: for each sample in the otter\_mimicit\_cgd dataset, we look for llava and llava\_dial samples that use the same image, and then build a new sample in a "llava/llava\_dial conversations then otter\_mimicit\_cgd conversation" fashion (as shown in \tref{tab:data-summary}).

\subsection{Results}
\subsubsection{Comparison with \qwen and \sparkles}
\paragraph{Experimental Setting}

We utilize the datasets (1) to (8) as illustrated in \tref{tab:data-summary}. We deliberately exclude the dialogue data from \sparkles to thoroughly investigate whether our newly proposed data blending technique can facilitate the interweaving of multi-image multi-round chats without the incorporation of new data. \llama-13B is employed as our language model, and the model is trained over 6 epochs. All other settings remain consistent with those outlined in \sref{sec:result_of_mrsi}.
\begin{figure}[t]
    \centering
     \includegraphics[width=1.0\textwidth]{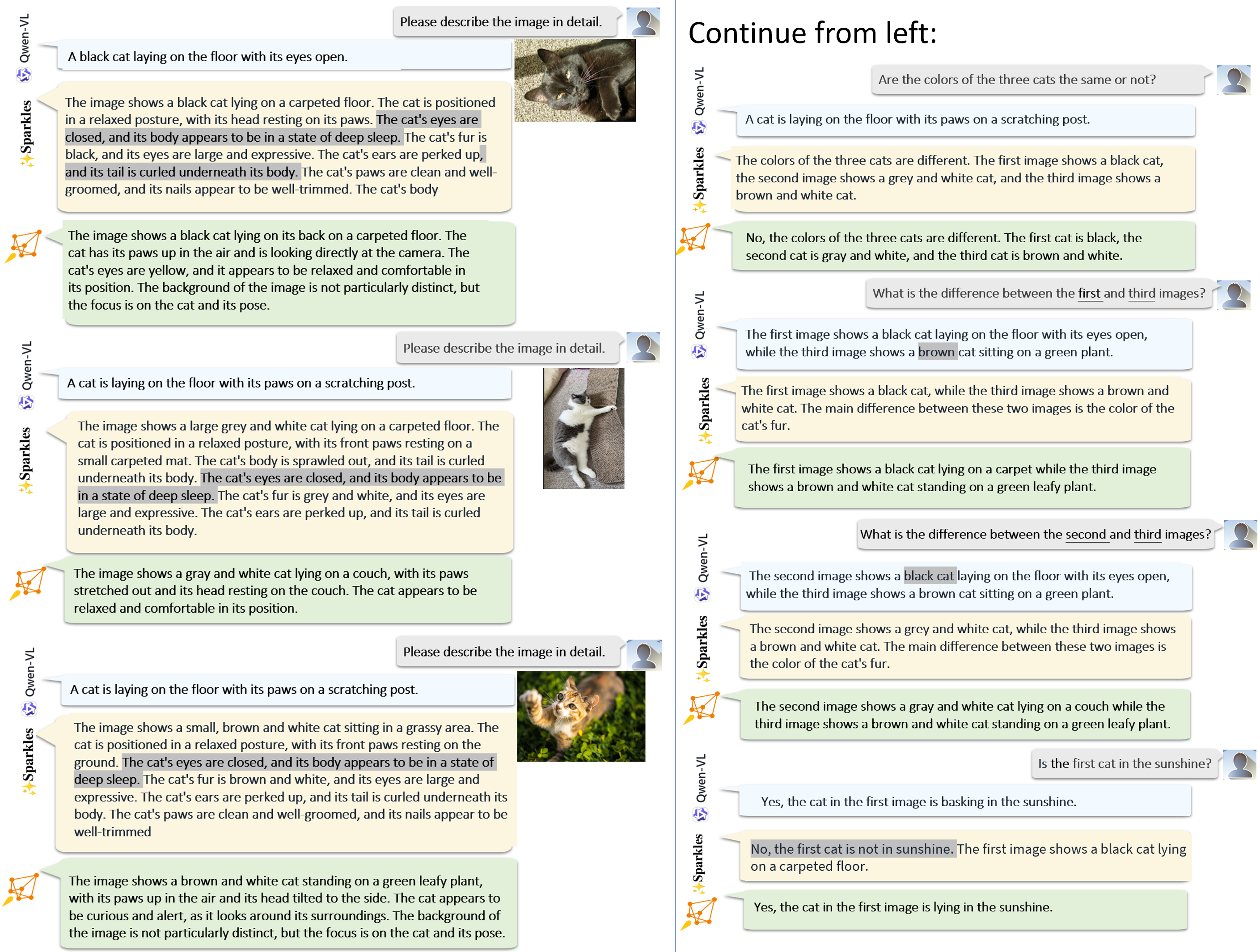}
    \caption{
The above example (the conversation is from left to right panel) illustrates the difference among \ours-13B, \qwen, and \sparkles. 
\qwen provides considerable short and accurate answers on describing the individual image but fails to remember the first image at the last second question. 
While \sparkles is good at interpreting the difference between images but provides inaccurate information when describing individual images. 
}
\label{tab:compare1}
\end{figure}
\begin{figure}
    \centering
     \includegraphics[width=1.02\textwidth]{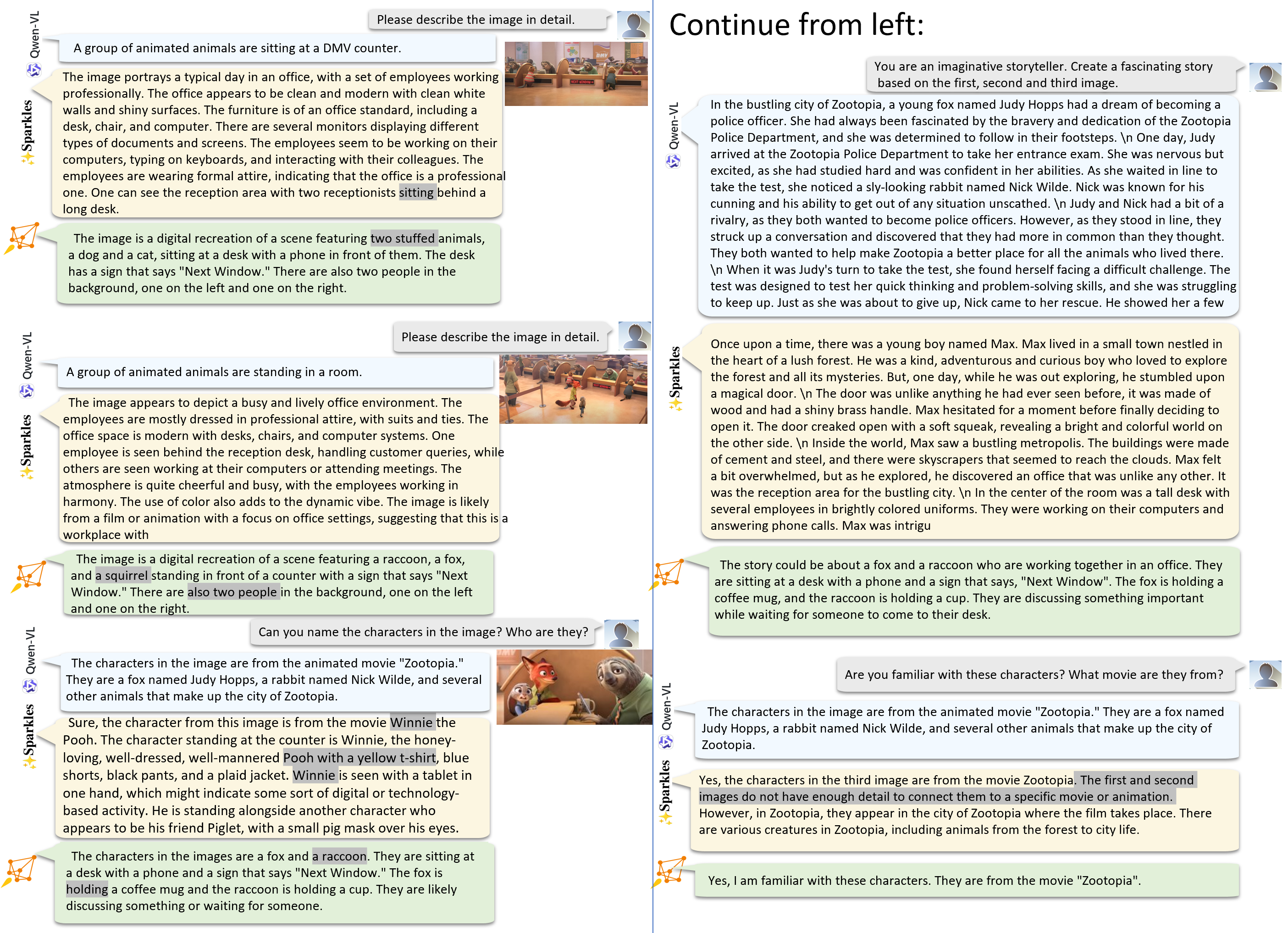}
\caption{
The given example, displayed in a left-to-right panel conversation format, demonstrates the disparities among \ours-13B, \qwen, and \sparkles. 
\qwen excels in delivering succinct and accurate responses when describing individual images. Conversely, \sparkles elaborates on the first two images but inaccurately identifies the scene as being from "Winnie" although later correctly attributing it to "Zootopia". 
When it comes to narrative skills, both \qwen and \sparkles exhibit proficiency in story-telling, a skill that our model lacks. 
This deficiency can be attributed to the absence of narrative-centric content within the training data utilized for this model. 
In~\fref{tab:compare}, we will compare the model trained with and without narrative-centric content.}
\label{tab:compare2}
\end{figure}




\paragraph{Demo Results}
We compare \ours-13B with \qwen and \sparkles as illustrated in \fref{tab:compare1} and \fref{tab:compare2}. The tasks presented in \fref{tab:compare1} are unseen to all the trained models. Notably, \ours-13B outperforms in terms of answer quality when compared to the other models. Specifically, while \qwen excels at offering succinct and accurate descriptions of individual images, it struggles to recall the first or second images during subsequent questions. On the other hand, \sparkles excels at discerning differences between images, yet occasionally provides imprecise descriptions of individual images.

The tasks in \fref{tab:compare2} center around narratives. Narratives are the primary training focus of \sparkles and might be a part of \qwen's training data (as its data is proprietary), but they were not part of \ours's training (i.e., datasets (1) to (8) as mentioned in \tref{tab:data-summary}). Despite this, \ours continues to provide commendable descriptions of individual images and exhibits some narrative skills. In contrast, both \qwen and \sparkles demonstrate superior narrative abilities. Nonetheless, each model has its own set of limitations for specific questions.

It is worth noting that the training expenditure for \ours is significantly lower than that for \qwen and \sparkles, with the latter having utilized the pre-training checkpoint of MiniGPT4.

\subsubsection{Result of \ours}
\label{sec:small_scale}

\begin{figure}
    \centering
     \includegraphics[width=1.0\textwidth]{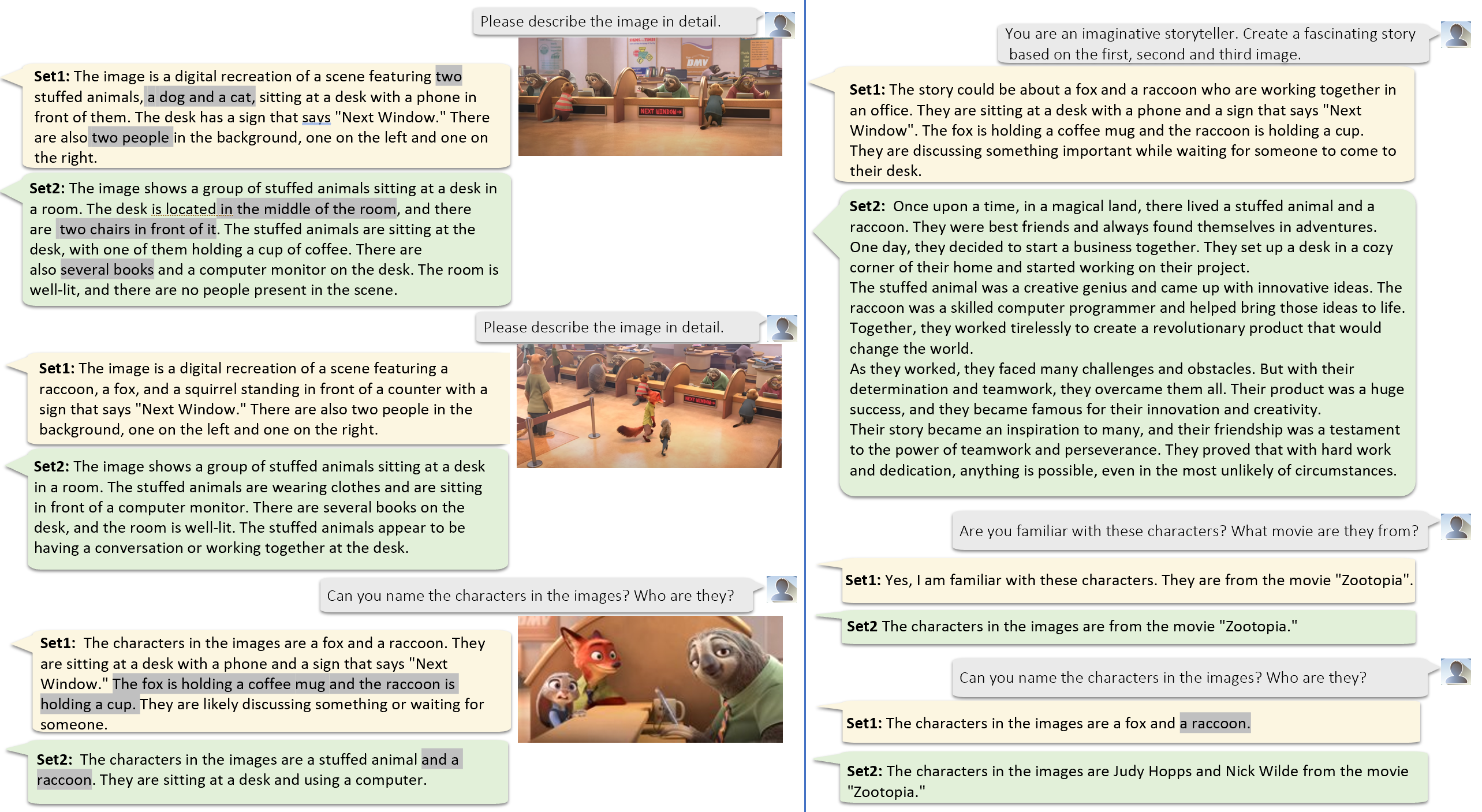}
\caption{
The above example (the conversation is from left to right panel) illustrates the differences between \ours-13B-Set1 and \ours-13b-Set2. We see that Set1 is better in describing the individual images (for example, Set1 can recognize the fox and "next window") but do not have ability to tell a story. 
While Set2 lost some ability to focus on the details of the images but can tell a story based on the given images.
}
\label{tab:compare}
\end{figure}
\paragraph{Experimental Setting}
The setting remains the same as mentioned above, with the addition of 6.5K more examples from \sparkles.

\paragraph{Demo result}




We perform comparisons between \ours-13B with and without incorporating data from \sparkles. For clarity, we will refer to \ours-13B without (and with) \sparkles's data as \ours-13B-Set1 (-Set2). First and foremost, by integrating \sparkles's data, \ours demonstrates enhanced narrative capability. Similar to \sparkles, the newly trained model also displays a reduced ability to concentrate on the details of individual images.

\begin{figure}
    \centering
     \includegraphics[width=1.0\textwidth]{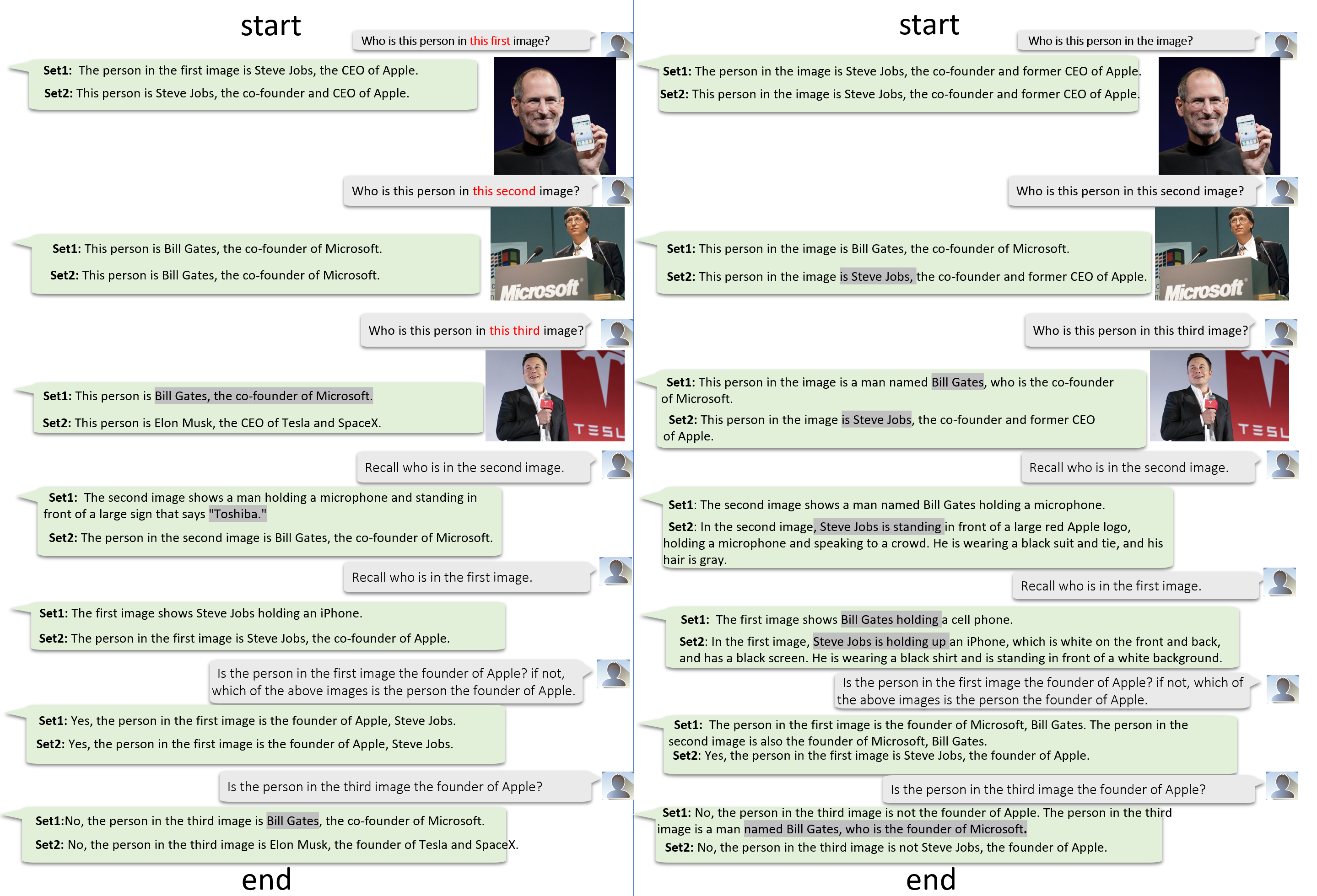}
\caption{
The given example, displayed in a left-to-right panel conversation format, demonstrates the disparities between \ours-13B-Set1 and \ours-13B-Set2, given a slightly different prompt series.}
\label{tab:compare5}
\end{figure}

Beyond the aforementioned, the introduction of additional data yields another intriguing observation. \ours-13B-Set2 exhibits increased sensitivity to prompt tuning compared to its predecessor. Specifically, as shown in \fref{tab:compare5}, a slight alteration to the prompt (highlighted in \red{red} text) without changing the question's meaning, leads \ours-13B-Set2 to provide disparate answers. Conversely, the original \ours-13B-Set1 tends to offer more congruent responses. We hypothesize that this heightened sensitivity results from an imbalance in question formats/templates introduced by \sparkles's dataset.

For cross-comparison between \ours and \qwen/\sparkles, please refer to \fref{tab:compare3} and \fref{tab:compare4}.


\subsubsection{\ours with \llama-70B}
We have initiated training with \llama-70B, maintaining the same training settings as outlined in~\sref{sec:small_scale}. However, the resulting model is not adequately trained. We conjecture that the hyper-parameters optimal for \llama-13B may not be suitable for \llama-70B; for instance, the learning rate might be excessive, and the number of training epochs insufficient. Perfecting the training of \ours-70B is earmarked for future work.

\subsubsection{Other Learning}

\begin{itemize}
    \item \textbf{Exploration of Projection Layers:} We experimented with two different projection layers to bridge visual encoders and \llms: a single linear layer and a Vision Transformer layer. We did not observe any benefits from the Vision Transformer approach in the preliminary phase, so we decided not to pursue this route further.
    \item \textbf{Advanced Data Blending Techniques:} We explored more intricate data blending methods, such as shuffling the image ID of the \otter and \llava datasets. For example, in the \otter dataset, the paired images were later referenced as the first and third images by inserting another image as the second one. However, our experiments led to deteriorated performance, characterized by incomplete sentences and incorrect references. Upon reviewing the data, we hypothesized that these issues were probably due to incorrect references in the training data during the data blending process.
\end{itemize}

\section{Limitations and Conclusions}
\label{sec:conclusions}
\paragraph{Limitations}
Given that the focus of this work is not on benchmarking evaluations, we abstained from incorporating any such results. This might have resulted in the demonstrations illustrated in the paper appearing biased and not being comprehensive. Additionally, we have observed that data is a pivotal component to achieve high-quality \lvlms, but we were unable to provide such datasets due to constraints on resources. We acknowledge that larger language models can potentially offer superior model quality, but we encountered difficulties in training a model based on \llama-70B. Attempts were made to train with the \llama-70B model, but the end results were suboptimal. We also noted the hallucination issue with \ours, a problem inherited from the \llama family.

\paragraph{Conclusions}
 In conclusion, We propose  \ours, the Multi Round and Multi Images Chat framework, a pioneering solution offering open-source support for multi-image dialogues and featuring an innovative multi-modal casual attention mechanism and efficient data blending techniques. Our approach surpasses contemporaneous models in scalability enabling enhanced adaptability in diverse interactive scenarios, without incurring additional training costs or complexity. This breakthrough facilitates unprecedented advancements in large vision and language models, laying a robust foundation for the seamless integration of visual and textual information in future multi-modal models.

\section*{Contributions}
\label{sec:contribution}

\noindent \textbf{ZY}: Full engagement and project lead. 
\noindent \textbf{XW}: Training/evaluation pipeline development and writing.
\noindent \textbf{CL}: Data support.
\noindent \textbf{MZ}: Training pipeline support.
\noindent \textbf{QH}: DeepSpeed ZeRO feature adaptation.
\noindent \textbf{OR}: DeepSpeed ZeRO feature adaptation.
\noindent \textbf{AAA}: Software support.
\noindent \textbf{SR}: Consulting.
\noindent \textbf{YH}: Team lead.

\section*{Acknowledgment}
We thank the entire DeepSpeed team for their support.{
 \bibliographystyle{plain}
\bibliography{ref.bib}
}

\appendix

\counterwithin{figure}{section}
\counterwithin{table}{section}

\section{More examples}

\begin{figure}
    \centering
     \includegraphics[width=1.0\textwidth]{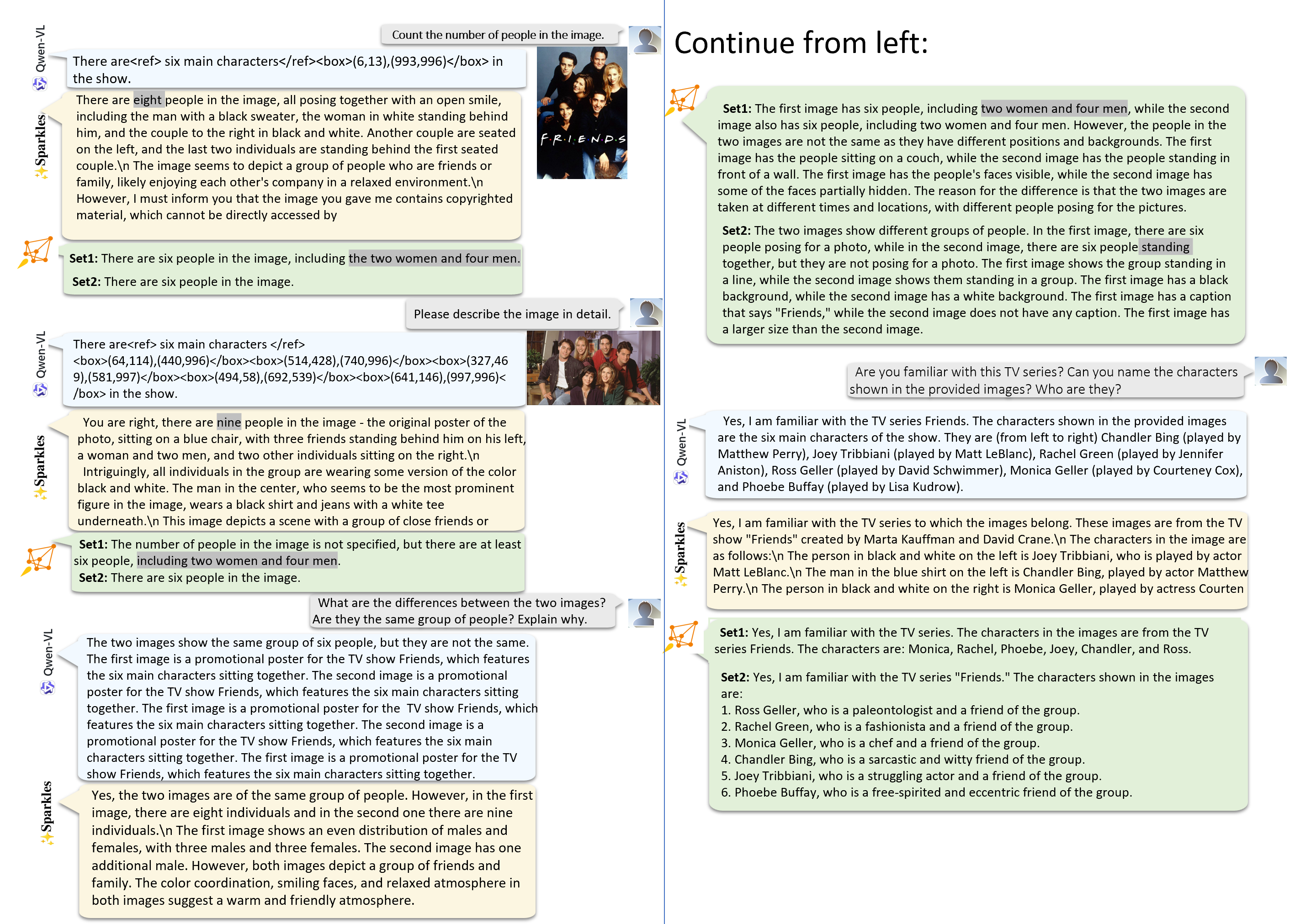}
\caption{
The given example, displayed in a left-to-right panel conversation format, demonstrates the disparities among \ours-13B-Set1 and \ours-13B-Set2, \qwen, and \sparkles.}
\label{tab:compare3}
\end{figure}

\begin{figure}
    \centering
     \includegraphics[width=1.0\textwidth]{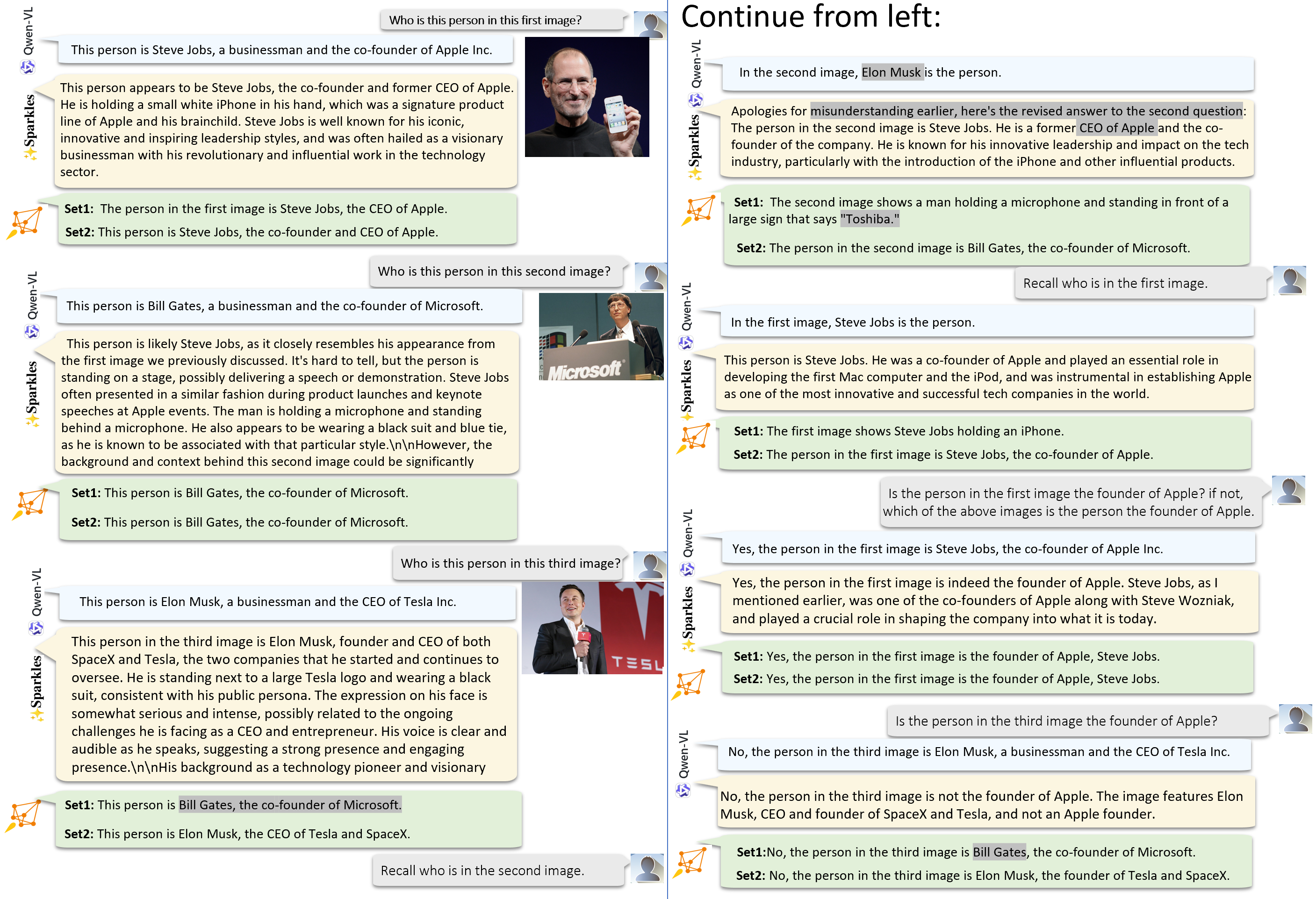}
\caption{
The given example, displayed in a left-to-right panel conversation format, demonstrates the disparities among \ours-13B-Set1 and \ours-13B-Set2, \qwen, and \sparkles.}
\label{tab:compare4}
\end{figure}

\end{document}